%% file: main.tex
\documentclass[10pt,twocolumn,letterpaper]{article}

\usepackage[pagenumbers]{iccv} % To force page numbers, e.g. for an arXiv version

\input{preamble}

\definecolor{iccvblue}{rgb}{0.21,0.49,0.74}
\usepackage[pagebackref,breaklinks,colorlinks,allcolors=iccvblue]{hyperref}

\usepackage[accsupp]{axessibility}

\def\paperID{4721} % *** Enter the Paper ID here
\def\confName{ICCV}
\def\confYear{2025}

\title{ChartPoint: Guiding MLLMs with Grounding Reflection for Chart Reasoning}

\author{Zhengzhuo Xu$^{1,2}$\thanks{$*$: Equal contributions. $\dagger$: Corresponding authors.} ~~ SiNan Du$^{1*}$ ~~ Yiyan Qi$^{2\dagger}$ ~~ SiwenLu$^3$ ~~ Chengjin Xu$^2$  ~~ Chun Yuan$^{1\dagger}$ ~~ Jian Guo$^{2,4}$ \\
$^1$Tsinghua University ~~ $^2$International Digital Economy Academy ~~ $^3$Beihang University \\
$^4$Hong Kong University of Science and Technology (Guangzhou)
}

\begin{document}
\maketitle
\input{sec/0_abstract}

\input{sec/1_intro}

\input{sec/2_related_work}

\input{sec/3_method}

\input{sec/4_experiment}

\input{sec/5_further_discussion}
\input{sec/6_conclude}
\clearpage

\section*{Acknowledgement}
This work was supported by the National Key R\&D Program of China (2022YFB4701400/4701402), SSTIC Grant \\(KJZD20230923115106012, KJZD20230923114916032, GJHZ20240218113604008), and Beijing Key Lab of Networked Multimedia.

{
\small
\bibliographystyle{ieeenat_fullname}
\bibliography{main}
}

\clearpage
\input{sec/X_suppl}

\end{document}

%% file: preamble.tex
\usepackage{booktabs}
\usepackage{multirow}
\usepackage{overpic}
\usepackage[table]{xcolor}
\usepackage{tcolorbox}
\usepackage{listings}
\definecolor{darkNavy}{RGB}{34, 36, 73}
\definecolor{lightBlue}{RGB}{156, 179, 212}
\definecolor{dustyRose}{RGB}{175, 90, 118}
\tcbuselibrary{listings}
\tcbuselibrary{breakable}

\definecolor{ptred}{RGB}{248, 207, 218}
\definecolor{gtgreen}{RGB}{217, 233, 206}

\definecolor{ao(english)}{rgb}{0.0, 0.5, 0.0}
\definecolor{deepred}{RGB}{177, 29, 5}
\newcommand{\good}[1]{\textcolor{ao(english)}{{\small \textbf{#1}}}}
\newcommand{\bad}[1]{\textcolor{deepred}{\small \textbf{#1}}}

\newcommand{\myparagraph}[1]{\vspace{1mm}\noindent{{\textbf{\textit{#1}}}}}
\newcommand{\Model}{ChartPoint}
\newcommand{\Dataset}{\texttt{ChartPoint-SFT-62k}}
\newcommand{\Mone}{ChartPoint$_\text{Q2}$}
\newcommand{\Mtwo}{ChartPoint$_\text{Q2.5}$}

%% file: sec/0_abstract.tex
\begin{abstract}
Multimodal Large Language Models (MLLMs) have emerged as powerful tools for chart comprehension. However, they heavily rely on extracted content via OCR, which leads to numerical hallucinations when chart textual annotations are sparse. While existing methods focus on scaling instructions, they fail to address the fundamental challenge, i.e., reasoning with visual perception. In this paper, we identify a critical observation: MLLMs exhibit weak grounding in chart elements and proportional relationships, as evidenced by their inability to localize key positions to match their reasoning. To bridge this gap, we propose PointCoT, which integrates reflective interaction into chain-of-thought reasoning in charts. By prompting MLLMs to generate bounding boxes and re-render charts based on location annotations, we establish connections between textual reasoning steps and visual grounding regions. We further introduce an automated pipeline to construct {\Dataset}, a dataset featuring 19.2K high-quality chart samples with step-by-step CoT, bounding box, and re-rendered visualizations. Leveraging this data, we develop two instruction-tuned models, {\Mone} and {\Mtwo}, which outperform state-of-the-art across several chart benchmarks, e.g., +5.04\% on ChartBench.
\end{abstract}
\vspace{-20pt}

%% file: sec/1_intro.tex
\section{Introduction}
\label{sec_intro}

Recently, with Large Language Models (LLMs) demonstrating strong understanding and generalization capabilities~\cite{GPT-1,qwen,deepseekllm,llama}, Multimodal Large Language Models (MLLMs) have become the mainstream for processing multimedia data such as images and videos~\cite{gpt4o, qwen-vl, deepseekvl, llava, ma2024followyouremoji}. Charts, as an intuitive way to present complex data, are widely adopted in documents and on the internet. However, current MLLMs heavily rely on optical character recognition (OCR) results when processing charts. When the text information extracted by OCR is limited, the MLLMs struggle to interpret the charts accurately, even leading to numerical hallucinations~\cite{MMC, ChartBench, charxiv}. Thus, extracting chart content accurately and attaining profound chart comprehension continues to be challenging tasks.

Existing methods attempt to address this issue through Supervised Fine-Tuning (SFT), including using more instruction-tuning data~\cite{Chartllama, MMC, chartgemma}, increasing the chart resolution~\cite{TinyChart}, or adopting more meticulously crafted alignment training techniques~\cite{ChartAst, ChartReformer, ChartMoE}. However, MLLMs still exhibit a limited perception of chart content. Recently, the inference-time scaling law and the reasoning models trained on it have exhibited impressive and in-depth reasoning capabilities~\cite{OpenAIO1, DeepseekR1}. Chain-of-Thought (CoT) training has notably enhanced LLMs' proficiency in mathematics, logic, and code~\cite{CoT, MCoT}. This motivates us to refine reasoning paradigms and inference formats of MLLMs on charts, especially in scenarios with sparse text annotations.

\input{figs/texs/teaser}

Do current MLLMs truly grasp the correct logic for chart interpretation? As depicted in Fig.~\ref{fig_teaser}, while the MLLMs present reasonable steps for chart-reading, the numbers they extract still contain significant errors. This situation prompts a crucial question: do MLLMs rely excessively on the extracted numbers when interpreting charts, thus lacking the capacity to read from chart elements and proportional relationships? To explore this, we employ the MLLMs~\cite{Qwen2VL, Qwen25VL} with satisfying localization capabilities, which can denote object positions using bounding boxes (BBox) or points. We prompt the model to point out the positions that match each reasoning step. Regrettably, MLLMs either overlook this request or generate entirely irrelevant positions. This implies that while the CoT approach bolsters the MLLM's logical processing based on numbers, it fails to enhance the model's fundamental numerical perception. Although CoT generates more inference tokens, it fails to enable additional interactions with chart or visual tokens, leading to limited perceptual improvement of MLLMs~\cite{visualcot, MVoT}. Hence, we enhance CoT by incorporating a reflective interaction process, where the model outputs BBoxes and engages with re-rendered charts (Fig.~\ref{fig_teaser}). Hence, we construct CoT data with BBox reflection called PointCoT. We enhance the model’s reasoning chain through a structured inference process and introduce an automated annotation pipeline leveraging chart-code pairs and advanced LLMs for precise step decomposition and key position localization. 
    
This pipeline consists of four stages. 1) \textit{Step Decomposition}: We collect high-quality chart-code pairs and use LLMs to generate a numerical question and corresponding CoT reasoning steps. The LLM labels each step as Grounding (requiring chart data extraction) or Reasoning. We will add point markers on the chart for all grounding steps. 2) \textit{Code Editing}: LLMs modify the code for all grounding steps by inserting special characters at key positions for easier position extraction. Directly employing MLLMs is unreliable for this task. Hence, we employ LLM-based code editing to achieve high success. Thus, each grounding step has a corresponding edited code. 3) \textit{Code Rendering}: We execute all modified code and re-render the charts. If any CoT step fails or triggers warnings, we discard the sample. 4) \textit{Position Localization}: We perform OCR on each rendered chart to extract embedded character positions. Through format checks, we ultimately derive BBoxes for grounding steps. Ultimately, we construct 19.2K samples, each containing a detailed CoT process and position annotations. We further present {\Dataset}, a dataset of 62.3K instructions, along with two SFT models called {\Mone} and {\Mtwo}. We achieve significant improvements across chart benchmarks, demonstrating the effectiveness of PointCoT. Our contributions are:
\setlength{\leftmargini}{20pt}
\begin{enumerate}[itemsep=0pt, topsep=0pt, label=\alph*)]
    \item We introduce PointCoT, which enables the MLLM to verify whether its reasoning steps align with the chart content using generated bounding boxes.
    \item We present {\Dataset}, a dataset containing 63.2K instruction-tuning samples. We also provide a data annotation pipeline to label the corresponding chart locations for CoT steps. 
    \item We propose the {\Mone} and {\Mtwo} based on proposed instruction data. Extensive experiments demonstrate that our models achieve state-of-the-art performance in chart understanding benchmarks.
\end{enumerate}

%% file: figs/texs/teaser.tex
\begin{figure}[t!]
    \centering
      \begin{overpic}[width=1\linewidth, grid=False]{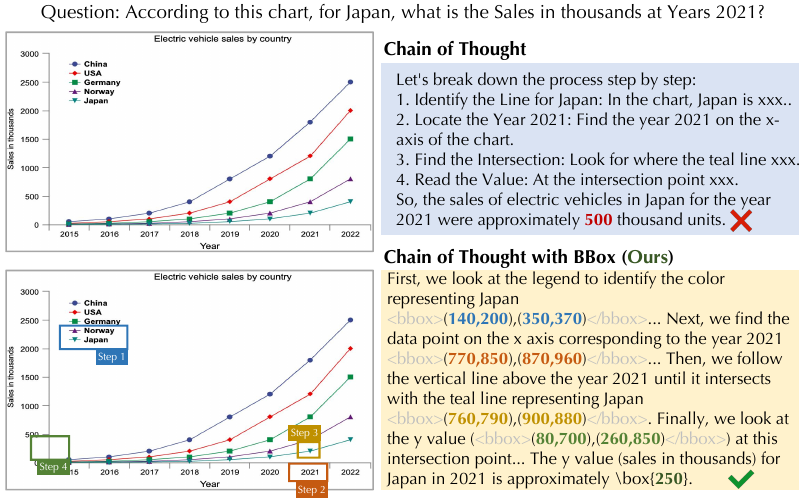}
      \end{overpic}
      \captionsetup{skip=3pt}   
      \caption{Comparison between vanilla CoT and proposed CoT with bounding box reflection on Qwen2-VL~\cite{Qwen2VL}. Vanilla CoT fails to introduce visual-level reflections. We re-render the generated BBox on the query chart to verify area focus and successfully improve the precision of the extracted numbers.}
      \vspace{-18pt}
    \label{fig_teaser}
\end{figure}

%% file: sec/2_related_work.tex
\section{Related Works}
\label{sec_related}

\myparagraph{Multimodal Large Language Models} adopt projectors to connect LLMs with visual encoders to understand images and demonstrate remarkable performance~\cite{Connector-S}. Some works employ QFormers~\cite{BLIP2} for modal alignment on large image-text pair datasets~\cite{BLIP2, Flamingo, Mplug, qwen-vl}. Other works further simplify the architecture with a linear projector and extend the instruction tuning paradigm to visual tasks~\cite{llava, BLIP3}. Training strategies and data quality are crucial for the development of MLLMs. The GPT series~\cite{GPT-1,GPT-3,GPT4, OpenAIO1} and Claude series~\cite{claude3} are the models with SOTA performance. The LLaMA series~\cite{llama, llama-2, llama3, llama3herdmodels} initially leads the open-source community and spawns works like the LLaVA series~\cite{llava,llava15,llava16}. The Qwen series~\cite{qwen, qwen2, qwen25, qwen-vl, Qwen2VL, Qwen25VL} and Intern series~\cite{intern,internlm2,internlm-xcomposer, internlm-xcomposerv2, internvl1, internvl15, InternVL25} have further elevated the performance of open-source models to SOTA level. The DeepSeek series~\cite{deepseekmoe, deepseekllm, deepseekv2, deepseekv3, deepseekvl, deepseekvl2, DeepseekR1} and Mistral series~\cite{mistral7b} conduct in-depth explorations of the Mixture of Experts architecture for MLLMs.

\myparagraph{Chart Reasoning} involves using MLLMs for tasks like question answering, description, analysis, and summarization of charts. Two-stage methods center on generating intermediate chart representations via specialized extraction modules. These representations can take forms such as markdown, as explored in \cite{Pix2Str, MatCha, DePlot}, or dictionaries, as seen in \cite{OneChart, ChartVLM}. Subsequently, they are supplied as text prompts to LLMs. End-to-end methods attempt to optimize MLLMs with more chart-related instructions~\cite{Chartllama, chartadapter}. Alignment training is employed to supplement prior knowledge in the chart domain, e.g., tabular~\cite{ChartPaLI,ChartAst,MMC}, markdown~\cite{Ureader,mPLUG2}, JSON~\cite{ChartReformer} or dictionaries~\cite{vprochart}. ChartThinker~\cite{ChartThinker} and DOMINO~\cite{DOMINO} propose the CoT for chart reasoning, and LaMenDa~\cite{LAMENDA} further integrates step-by-step reasoning into the supervised fine-tuning stage. TinyChart~\cite{TinyChart} upsamples the chart resolution and achieves a notable performance improvement. Moreover, recent works~\cite{askchart, ChartMoE} attempt to combine the advantages of the above approaches using the mixture of experts architecture. 

\myparagraph{Multimodal Chain of Thought} aims to extend text-based CoT reasoning~\cite{CoT, DeepseekR1} to multimodal scenarios to enhance performance in tasks requiring logical reasoning. Some two-stage works either convert visual information into text~\cite{MCoT, CCoT, Kam-cot} or sample key image information (e.g., region~\cite{visualcot} or coordinate~\cite{scaffolding}). GoT~\cite{GoT-CQA} generates directed acyclic graphs to assist reasoning. Recently, structured reasoning is proposed to enhance the robustness of the CoT. Both InsightV~\cite{Insight-v} and LLaVA-CoT~\cite{Llava-cot} propose a reasoning framework based on human design to solve a wide range of visual question-answering problems. Further research aims to enhance the interaction between the reasoning steps and the query image in structured scenarios~\cite{MVoT, Heima}.

%% file: sec/3_method.tex
\section{Proposed Method}

\subsection{PointCoT}
To enhance the reasoning process, we focus on constructing extensive thinking-chain data for chart-based Q\&A while leveraging coordinate points to guide the model's attention to relevant chart regions. To ensure the model learns correct chart-reading logic, we select charts without datapoint annotations, preventing it from extracting answers directly via OCR. Specifically, our metadata construction is based on the ChartAlign dataset~\cite{ChartMoE}, which comprises one million quadruples (table, JSON, code, chart) sourced from ChartQA~\cite{ChartQA}, PlotQA~\cite{PlotQA}, and ChartY~\cite{OneChart}. Our objective is to generate chain-of-thought reasoning data for charts and incorporate coordinate-based cues at each step to justify the model's focus region. The following sections detail the step-by-step process of constructing the training data.

\input{figs/texs/pipeline_cot}

\subsection{Construction of Structured Reasoning}
Researchers typically employ advanced LLMs to decompose and expand the reasoning process of the text data, aiming to obtain long chain-of-thought inference processes. Recent studies have also demonstrated that distillation learning based on such data enables smaller models to acquire strong reasoning abilities~\cite{DeepseekR1}. Unlike general visual Q\&A tasks that require diverse knowledge and reasoning styles, chart Q\&A exhibits a structured thought process, i.e., the model infers correct numbers from visual elements like legends and coordinate systems through consistent steps, which can be enhanced with structured reasoning training. Fig.~\ref{fig_pipeline_cot} elaborately outlines the process of our reasoning data construction. Our primary focus lies in straightforward chart comprehension, centered around chart data points Q\&A. Although the reasoning process appears structured, this structure does not arise from artificial constraints. Instead, it emerges naturally from the inherent logic of chart reading, imparting a degree of structural consistency to the decomposed CoT steps.

Fig.~\ref{fig_pipeline_cot} presents an example chart and the generated JSON. First, we utilize the teacher model (i.e., Qwen2.5-72B~\cite{qwen25}) to pose a datapoint-related question based on the plotting code. We require the teacher to provide a step-by-step reasoning process and the final answer. We employ few-shot examples to standardize the step-decomposition format and ask the teacher model to classify each sub-step into two categories: \textit{Grounding} and \textit{Reasoning}. Refer to Appendix~\ref{apdx_prompt} for the detailed prompt. Grounding steps focus on identifying the positions of chart elements, such as locating points on the axes or entries in the legend. Reasoning steps involve making logical inferences based on information obtained from previous grounding steps. This classification helps incorporate specific bounding boxes for steps that require element localization, thereby offering precise positional guidance. Finally, we instruct the teacher model to generate outputs in JSON format. Samples that pass both the format validation and key integrity checks proceed to the following processing stage. 

\input{figs/texs/pipeline_code_edit}

\subsection{Construction of Point Annotation}
Our goal is to incorporate location supervision into all grounding steps, guiding the model to follow human-like chart-reading logic. We believe the generated bounding boxes not only validate the grounding steps but also encourage the model to re-examine the original input chart. Therefore, we implement point-based CoT training through grounding. MVoT~\cite{MVoT} also achieves similar observations in other structured scenarios, e.g., puzzle-solving games.

Fig.~\ref{fig_pipeline_code_edit} elaborately depicts how we add the position points to all grounding steps. Our modifications are based on revising the plotting code and OCR on the re-rendered chart. Specifically, we instruct the teacher model to identify the relevant elements (e.g., legend or title) or positions (e.g., datapoints or corresponding horizontal and vertical coordinates) for each grounding step. Next, the teacher model modifies the plotting code based on the identified positions by inserting a special symbol into the chart element text or marking a specific position using \texttt{plt.text()}. This insertion not only highlights key positions but also facilitates the quick detection of unique characters with OCR tools.

After passing the integrity check, the edited code is re-rendered to generate the updated chart. We then apply OCR to the re-rendered chart to extract the coordinates of the inserted special characters. To enhance extraction accuracy and success rates, we employ multiple OCR tools sequentially. A minimum width is defined for the bounding boxes generated by special characters, and any boxes more minor than the threshold are adjusted based on the center point and the pre-set width. Each grounding step is associated with an edited code, a re-rendered chart, and the detected positions from OCR. Refer to Appendix~\ref{apdx_demo} for details.

\input{figs/texs/pipeline_instruction}
\input{tables/tab_data_process}

\subsection{Construction of Instruction}
After obtaining the bounding boxes for all grounding steps, we begin constructing instruction data with location annotations. Fig~\ref{fig_pipeline_instruction} illustrates the process to construct {\Dataset}, which primarily includes four formats and 62K Q\&A pairs.

\textit{Type 1: Standard VQA.} The raw chart and question are used as input. 1) Supervised with ground truth answer. Unlike previous ChartQA~\cite{ChartQA}, the data points are not directly labeled with text, making the questions more challenging. 2) Supervised with CoT steps and the answer as long text supervision. Here, the bounding boxes from the grounding step are excluded to prevent potential data leakage and avoid affecting other formats. \textit{Type 2: Localization Task.} Different from direct Q\&A, we introduce intermediate steps into the query prompt. The ground truth is changed from the answer to the predicted bounding box, which is a localization task. \textit{Type 3: Reasoning with Edited Chart.} The bounding box annotations in the previous grounding steps will be redrawn on the vanilla chart to attract attention to the key position, aiding the model in learning the correct visual reasoning logic. If the next step is also a grounding step, the model will continue to predict the next bounding box based on the edited chart. \textit{Type 4: Reasoning Steps.} If the next step is the reasoning step, it will be added to the query prompt directly. Once the final step is processed, the supervised ground truth will be the final answer.

\input{figs/texs/dataset_statistic}
\input{tables/tab_sft_data}

\input{tables/tab_chartqa}

\subsection{Quality Control}
Considering the lengthy data generation process, we implement quality control at every step and track success rates. As shown in Tab.~\ref{tab_data_process}, we randomly sample 66.84k quadruples from ChartMoE-Align~\cite{ChartMoE}. 1) We expand the reasoning process based on the plot code and perform the integrity check on the generated JSON (Fig.~\ref{fig_pipeline_cot}). We employ GPT-4o~\cite{gpt4o} to review the generated Q\&A given the metatable data to filter out mismatched samples. The pass rate is 96.17\%. 2) We modify the plotting code by incorporating the grounding step as the instruction (Fig.~\ref{fig_pipeline_code_edit}). We ensure the code integrity and verify the presence of the required unique character in the code. The pass rate is 75.84\%. 3) We execute the modified code to render the edited charts. One case will be discarded if any code execution fails, resulting in a lower success rate of 51.04\%. 4) We use OCR to detect special characters and extract the bounding boxes. We discard the cases where OCR fails or detects multiple occurrences. This step achieves a success rate of 77.17\%. Finally, we construct 19.2K charts and 62.3K instruction data as illustrated in Fig.~\ref{fig_pipeline_instruction}. We randomly sample 100 cases, which are reviewed by at least three experts to evaluate the bounding box quality of the grounding step based on the process in Fig.~\ref{fig_pipeline_cot}. 91\% of the cases meet the desired standard.

\subsection{Statistics}
Fig.~\ref{fig_dataset_statistic} presents the statistics of {\Dataset}. As shown in Fig.~\ref{fig_dataset_statistic} (left), we carefully count all the CoT steps and organize the samples based on the length of the CoT steps.  Most samples contain 3-5 CoT steps. Notably, the grounding steps are typically longer (length $>$ 3) than the reasoning steps, which are predominantly short (length $\leq$ 3) and generally focus on summary-style analyses. This is because our Q\&A primarily addresses numerical data points without requiring complex numerical reasoning, allowing the dataset to effectively capture the essential \textit{visual logical} based more on grounding.  As shown in Fig.~\ref{fig_dataset_statistic} (right), we primarily focus on three chart types, i.e., line (33.6\%), pie (9.3\%), and bar (57.1\%) charts, which is consistent with the distribution of mainstream chart datasets~\cite{ChartQA, PlotQA}.

\subsection{\Model}
We integrate bounding box reflection into the inference. The baseline’s grounding ability is critical for instruction tuning. Hence, we select Qwen2-VL~\cite{Qwen2VL} and Qwen2.5-VL~\cite{Qwen25VL} as baselines due to their comprehensive grounding capabilities. They can be deployed based on LLaMA-Factory~\cite{llamafactory} to conduct convenient training. We perform a two-stage full fine-tuning process using the data in Tab.~\ref{tab_sft_data}. We utilize high-quality instruction data (including real-world annotated and diversely synthesized charts) for chart knowledge alignment to enhance the baseline’s performance. Then, we refresh the learning rate and conduct chart-specific annealing tuning in our PointCoT manner. The SFT models are named {\Mone} and {\Mtwo}, respectively.

%% file: figs/texs/pipeline_cot.tex
\begin{figure}[t!]
    \centering
      \begin{overpic}[width=1\linewidth, grid=False]{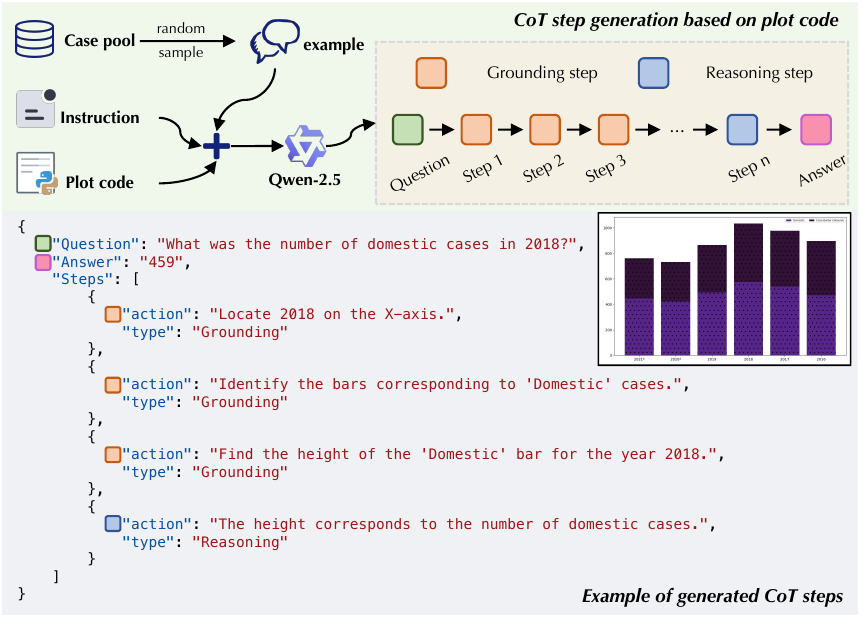}
      \end{overpic}
      \captionsetup{skip=3pt}      
      \caption{Chain of thought step generation based on plot code.}
      \vspace{-10pt}
    \label{fig_pipeline_cot}
\end{figure}

%% file: figs/texs/pipeline_code_edit.tex
\begin{figure}[t!]
    \centering
      \begin{overpic}[width=1\linewidth, grid=False]{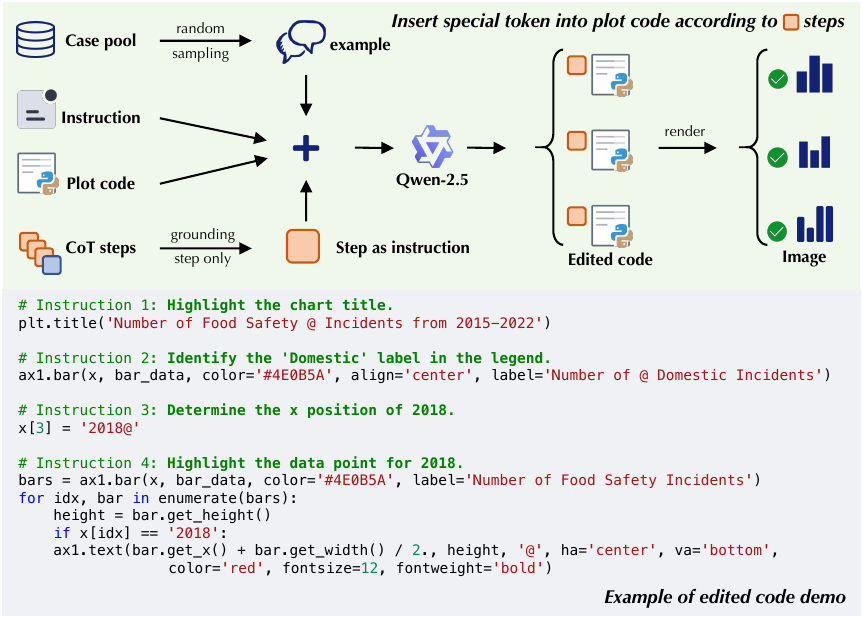}
      \end{overpic}
      \captionsetup{skip=3pt}      
      \caption{The pipeline of code editing with grounding steps.}
      \vspace{-10pt}
    \label{fig_pipeline_code_edit}
\end{figure}

%% file: figs/texs/pipeline_instruction.tex
\begin{figure}[t!]
    \centering
      \begin{overpic}[width=1\linewidth, grid=False]{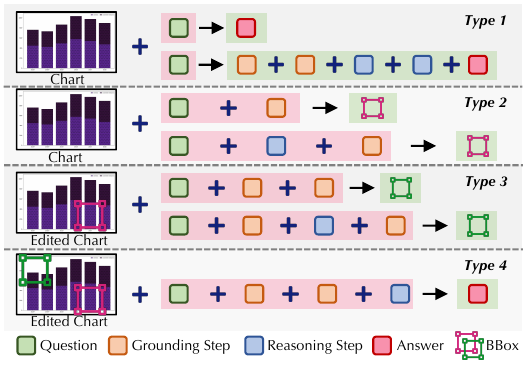}
      \end{overpic}
      \captionsetup{skip=3pt}      
      \caption{The process pipeline for constructing instruction data. The \colorbox{ptred}{red} / \colorbox{gtgreen}{green} indicates the instruction prompt / ground truth.}
      \vspace{-10pt}
    \label{fig_pipeline_instruction}
\end{figure}

%% file: tables/tab_data_process.tex
\begin{table}[t]
\centering
\captionsetup{skip=3pt}
\caption{Data processing steps and corresponding success rate. \# indicates the number of instructions.}
\resizebox{\linewidth}{!}{
\setlength{\tabcolsep}{5pt}
\begin{tabular}{@{}c|ccccc|c@{}}
\toprule[1pt]
Processing Step & Meta  & CoT   & Code  & Render & OCR   & QA\# \\ \midrule
Chart Number    & 66.84K & 64.28K & 48.75K & 24.88K  & 19.2K  & 62.3K \\
Success Rate    & -     & 96.17\% & 75.84\% & 51.04\%  & 77.17\% & -    \\ \bottomrule[1pt]
\end{tabular}
}
\label{tab_data_process}
\vspace{-18pt}
\end{table}

%% file: figs/texs/dataset_statistic.tex
\begin{figure}[t!]
    \centering
      \begin{overpic}[width=1\linewidth, grid=False]{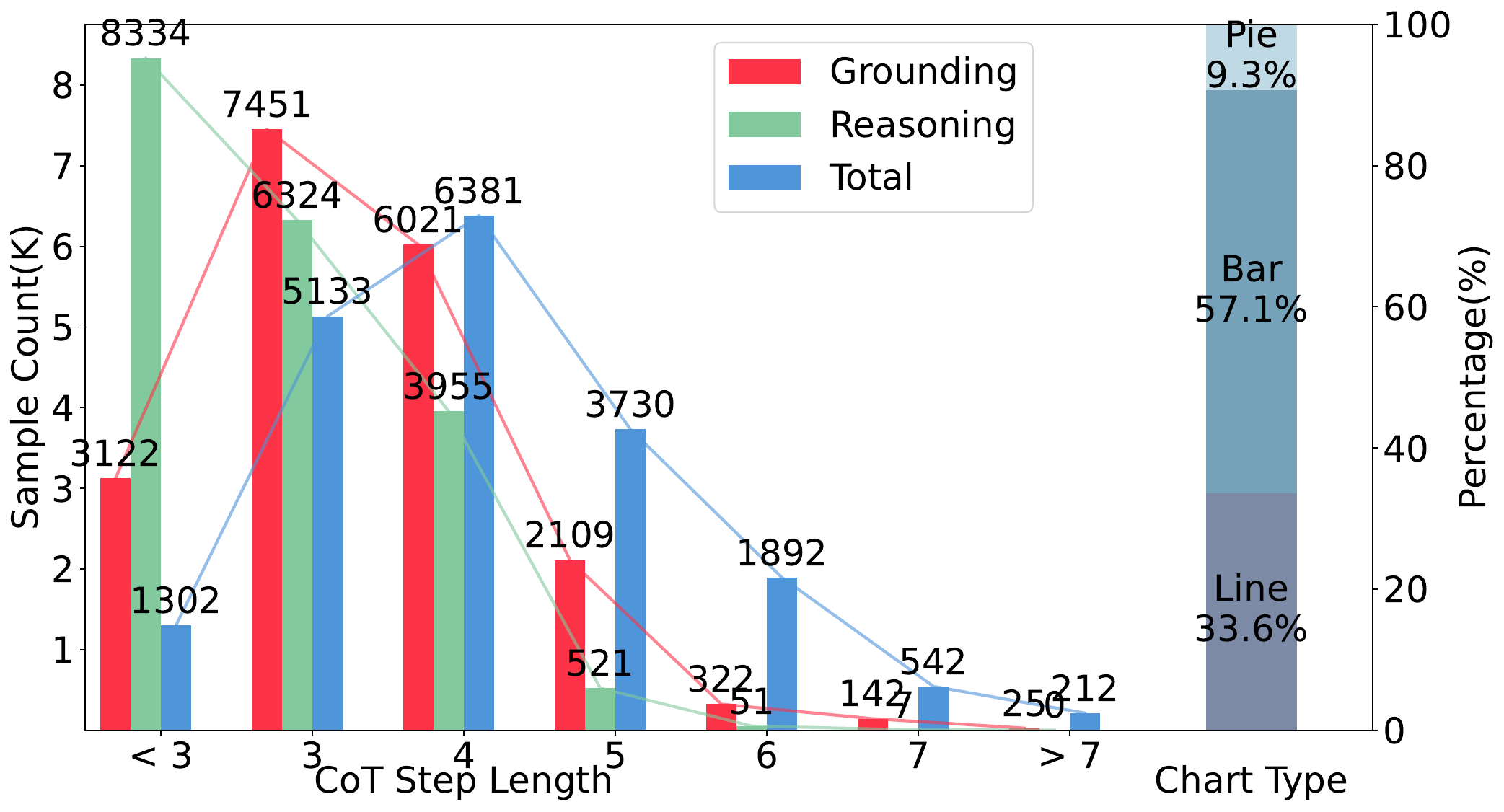}
      \end{overpic}
      \captionsetup{skip=3pt}      
      \caption{Statistic information of {\Dataset}. Left: Statistics on the number of CoT steps w.r.t. grounding, reasoning, and total steps. Right: chart type distribution.}
      \vspace{-10pt}
    \label{fig_dataset_statistic}
\end{figure}

%% file: tables/tab_sft_data.tex
\begin{table}[t]
\centering
\captionsetup{skip=3pt}
\caption{Instruction data used for {\Model} superivised training.}
\resizebox{\linewidth}{!}{
\setlength{\tabcolsep}{5pt}
\begin{tabular}{@{}ccc@{}}
\toprule[1pt]
\multicolumn{1}{c|}{Dataset}    & \multicolumn{1}{c|}{Description}                   & Number \\ \midrule
\multicolumn{3}{c}{\textit{Chart Knowledge Alignment Stage}}                                                 \\
\multicolumn{1}{c|}{MMC-Instruct~\cite{MMC}}        & \multicolumn{1}{c|}{VQA / Summariztion/ Reasoning} & 410K   \\
\multicolumn{1}{c|}{ChartGemma~\cite{chartgemma}} & \multicolumn{1}{c|}{VQA / Summariztion/ Reasoning} & 160K   \\
\multicolumn{1}{c|}{ChartQA~\cite{ChartQA}}    & \multicolumn{1}{c|}{VQA}                           & 28K    \\
\multicolumn{1}{c|}{ChartBench~\cite{ChartBench}} & \multicolumn{1}{c|}{VQA}                           & 30K    \\ \midrule
\multicolumn{3}{c}{\textit{Chart Specific Annealing Tuning Stage}}                                           \\
\multicolumn{1}{c|}{{\Dataset}} & \multicolumn{1}{c|}{VQA / Reasoning}               & 62K    \\ \bottomrule[1pt]
\end{tabular}
}
\label{tab_sft_data}
\vspace{-18pt}
\end{table}

%% file: tables/tab_chartqa.tex
\begin{table*}[t]
\centering
\captionsetup{skip=3pt}
\caption{The relaxed accuracy (\%) performance on \textbf{\textit{ChartQA}}. Ada.: Adaptive input resolution. Methods are sorted by relaxed average accuracy@0.05. All results are reproduced in the same inference manner by officially released model weights and prompts.}
\resizebox{\linewidth}{!}{
\setlength{\tabcolsep}{8pt}
\begin{tabular}{@{}ccccccccccccc@{}}
\toprule[1pt]
\multicolumn{1}{c|}{\multirow{2}{*}{Models}} & \multicolumn{1}{c|}{\multirow{2}{*}{Para.}} & \multicolumn{1}{c|}{\multirow{2}{*}{Baseline}} & \multicolumn{1}{c|}{\multirow{2}{*}{Res.}} & \multicolumn{3}{c|}{Relax Acc @0.05} & \multicolumn{3}{c|}{Relax Acc @0.10} & \multicolumn{3}{c}{Relax Acc @0.20} \\
\multicolumn{1}{c|}{} & \multicolumn{1}{c|}{} & \multicolumn{1}{c|}{} & \multicolumn{1}{c|}{} & Human & Aug. & \multicolumn{1}{c|}{Avg.} & Human & Aug. & \multicolumn{1}{c|}{Avg.} & Human & Aug. & Avg. \\ \midrule
\multicolumn{13}{c}{\textit{General MLLMs}} \\
\multicolumn{1}{c|}{LLaVA-v1.5~\cite{llava15}} & \multicolumn{1}{c|}{13B} & \multicolumn{1}{c|}{Vicuna~\cite{vicuna}} & \multicolumn{1}{c|}{@336} & 25.36 & 18.56 & \multicolumn{1}{c|}{21.96} & 28.56 & 23.52 & \multicolumn{1}{c|}{26.04} & 32.56 & 30.72 & 31.64 \\
\multicolumn{1}{c|}{Qwen-VL~\cite{qwen-vl}} & \multicolumn{1}{c|}{9.6B} & \multicolumn{1}{c|}{Qwen~\cite{qwen}} & \multicolumn{1}{c|}{@448} & 40.48 & 79.76 & \multicolumn{1}{c|}{60.12} & 43.20 & 82.56 & \multicolumn{1}{c|}{62.88} & 47.52 & 85.76 & 66.64 \\
\multicolumn{1}{c|}{Phi-3.5-Vision~\cite{phi3}} & \multicolumn{1}{c|}{4.2B} & \multicolumn{1}{c|}{Phi-3.5\cite{phi3}} & \multicolumn{1}{c|}{Ada.} & 60.08 & 83.52 & \multicolumn{1}{c|}{71.80} & 64.00 & 85.92 & \multicolumn{1}{c|}{74.96} & 68.16 & 89.36 & 78.76 \\
\multicolumn{1}{c|}{InternlmXC-v2~\cite{internlm-xcomposerv2}} & \multicolumn{1}{c|}{8B} & \multicolumn{1}{c|}{InternLM-v2~\cite{internlm2}} & \multicolumn{1}{c|}{@490} & 62.72 & 81.28 & \multicolumn{1}{c|}{72.00} & 66.72 & 84.08 & \multicolumn{1}{c|}{75.40} & 70.80 & 86.56 & 78.68 \\
\multicolumn{1}{c|}{InternVL-v2.5~\cite{InternVL25}} & \multicolumn{1}{c|}{8B} & \multicolumn{1}{c|}{InternLM-v2.5~\cite{internlm2}} & \multicolumn{1}{c|}{Ada.} & 65.44 & 86.48 & \multicolumn{1}{c|}{75.96} & 67.36 & 86.88 & \multicolumn{1}{c|}{77.12} & 68.80 & 87.44 & 78.12 \\
\multicolumn{1}{c|}{DeepSeekVL2~\cite{deepseekvl2}} & \multicolumn{1}{c|}{27B} & \multicolumn{1}{c|}{DeepSeek-v2~\cite{deepseekv2}} & \multicolumn{1}{c|}{@384} & 65.52 & 87.76 & \multicolumn{1}{c|}{76.64} & 67.52 & 88.08 & \multicolumn{1}{c|}{77.80} & 69.60 & 88.96 & 79.28 \\
\multicolumn{1}{c|}{Qwen2-VL~\cite{Qwen2VL}} & \multicolumn{1}{c|}{7B} & \multicolumn{1}{c|}{Qwen2~\cite{qwen2}} & \multicolumn{1}{c|}{Ada.} & 72.08 & \textbf{94.24} & \multicolumn{1}{c|}{83.16} & 75.76 & \textbf{94.72} & \multicolumn{1}{c|}{85.24} & 78.24 & \textbf{95.76} & 87.00 \\
\multicolumn{1}{c|}{Qwen2.5-VL~\cite{Qwen25VL}} & \multicolumn{1}{c|}{7B} & \multicolumn{1}{c|}{Qwen2.5~\cite{qwen25}} & \multicolumn{1}{c|}{Ada.} & \textbf{78.96} & 93.76 & \multicolumn{1}{c|}{\textbf{86.36}} & \textbf{81.12} & 94.16 & \multicolumn{1}{c|}{\textbf{87.64}} & \textbf{83.60} & 94.72 & \textbf{89.16} \\ \midrule
\multicolumn{13}{c}{\textit{Specialist Chart Models}} \\
\multicolumn{1}{c|}{Matcha~\cite{MatCha}} & \multicolumn{1}{c|}{282M} & \multicolumn{1}{c|}{Pix2Struct~\cite{Pix2Str}} & \multicolumn{1}{c|}{Ada.} & 37.12 & 86.64 & \multicolumn{1}{c|}{61.88} & 39.84 & 87.36 & \multicolumn{1}{c|}{63.60} & 43.52 & 88.56 & 66.04 \\
\multicolumn{1}{c|}{ChartVLM~\cite{ChartVLM}} & \multicolumn{1}{c|}{13B} & \multicolumn{1}{c|}{Vicuna~\cite{vicuna}} & \multicolumn{1}{c|}{Ada.} & 42.08 & 82.48 & \multicolumn{1}{c|}{62.28} & 43.84 & 82.88 & \multicolumn{1}{c|}{63.36} & 46.00 & 83.28 & 64.64 \\
\multicolumn{1}{c|}{DocOwl-v1.5~\cite{DocOwl}} & \multicolumn{1}{c|}{8B} & \multicolumn{1}{c|}{mPLUG-Owl2~\cite{mPLUG2}} & \multicolumn{1}{c|}{@448} & 47.44 & 91.52 & \multicolumn{1}{c|}{69.48} & 51.92 & 92.08 & \multicolumn{1}{c|}{72.00} & 56.72 & 93.12 & 74.92 \\
\multicolumn{1}{c|}{Deplot~\cite{DePlot}} & \multicolumn{1}{c|}{13.2B} & \multicolumn{1}{c|}{LLaVA-v1.6~\cite{llava16}} & \multicolumn{1}{c|}{Ada.} & 53.44 & 87.68 & \multicolumn{1}{c|}{70.56} & 56.80 & 88.48 & \multicolumn{1}{c|}{72.64} & 60.64 & 90.08 & 75.36 \\
\multicolumn{1}{c|}{OneChart~\cite{OneChart}} & \multicolumn{1}{c|}{13.3B} & \multicolumn{1}{c|}{LLaVA-v1.6~\cite{llava16}} & \multicolumn{1}{c|}{@1024} & 54.48 & 87.12 & \multicolumn{1}{c|}{70.80} & 57.60 & 87.84 & \multicolumn{1}{c|}{72.72} & 62.00 & 88.64 & 75.32 \\
\multicolumn{1}{c|}{ChartLlama~\cite{Chartllama}} & \multicolumn{1}{c|}{13B} & \multicolumn{1}{c|}{LLaVA-v1.5~\cite{llava15}} & \multicolumn{1}{c|}{@336} & 58.40 & 93.12 & \multicolumn{1}{c|}{75.76} & 61.20 & 93.60 & \multicolumn{1}{c|}{77.40} & 63.52 & 94.00 & 78.76 \\
\multicolumn{1}{c|}{ChartGemma+PoT~\cite{chartgemma}} & \multicolumn{1}{c|}{3B} & \multicolumn{1}{c|}{PaliGemma~\cite{paligemma}} & \multicolumn{1}{c|}{@448} & 67.84 & 85.28 & \multicolumn{1}{c|}{76.56} & 68.64 & 85.84 & \multicolumn{1}{c|}{77.24} & 69.84 & 86.32 & 78.08 \\
\multicolumn{1}{c|}{ChartAst~\cite{ChartAst}} & \multicolumn{1}{c|}{13B} & \multicolumn{1}{c|}{Sphinx~\cite{sphinx}} & \multicolumn{1}{c|}{@448} & 64.88 & 93.12 & \multicolumn{1}{c|}{79.00} & 66.24 & 93.84 & \multicolumn{1}{c|}{80.04} & 67.44 & 94.32 & 80.88 \\
\multicolumn{1}{c|}{TinyChart+PoT~\cite{TinyChart}} & \multicolumn{1}{c|}{3B} & \multicolumn{1}{c|}{TinyLlava~\cite{tinyllama}} & \multicolumn{1}{c|}{@768} & 70.24 & 90.72 & \multicolumn{1}{c|}{80.48} & 71.20 & 91.44 & \multicolumn{1}{c|}{81.32} & 72.40 & 92.56 & 82.48 \\
\multicolumn{1}{c|}{ChartMoE+PoT~\cite{ChartMoE}} & \multicolumn{1}{c|}{8B} & \multicolumn{1}{c|}{InternlmXC-v2~\cite{internlm-xcomposerv2}} & \multicolumn{1}{c|}{@490} & 78.32 & 90.96 & \multicolumn{1}{c|}{84.64} & 80.16 & 92.32 & \multicolumn{1}{c|}{86.24} & 82.08 & 93.60 & 87.84 \\
\rowcolor{blue!5}\multicolumn{1}{c|}{{\Mone}} & \multicolumn{1}{c|}{7B} & \multicolumn{1}{c|}{Qwen2-VL~\cite{Qwen2VL}} & \multicolumn{1}{c|}{Ada.} & 76.12 & \textbf{94.48} & \multicolumn{1}{c|}{85.28} & 78.36 & 94.96 & \multicolumn{1}{c|}{86.66} & 81.28 & 95.12 & 88.20 \\
\rowcolor{blue!5}\multicolumn{1}{c|}{{\Mtwo}} & \multicolumn{1}{c|}{7B} & \multicolumn{1}{c|}{Qwen2.5-VL~\cite{Qwen25VL}} & \multicolumn{1}{c|}{Ada.} & \textbf{81.36} & 94.12 & \multicolumn{1}{c|}{\textbf{87.74}} & \textbf{82.40} & \textbf{95.24} & \multicolumn{1}{c|}{\textbf{88.82}} & \textbf{84.48} & \textbf{95.76} & \textbf{90.12} \\ \bottomrule[1pt]
\end{tabular}
}
\label{tab_chartqa}
\vspace{-18pt}
\end{table*}

%% file: sec/4_experiment.tex
\section{Experiment}

\input{tables/tab_chartbench}

\subsection{Implement Details}
{\Model} is initialized from Qwen~\cite{Qwen2VL, Qwen25VL}, which employs a dynamic resolution input strategy. We keep all numerical coordinates within the range of $0-999$ to adapt to the tokenizer and the pretrain format of the coordinate system. We use LLaMA-Factory~\cite{llamafactory} for supervised fine-tuning over $2$ epochs. In the first 1\% of the training steps, we implement a warmup phase with a learning rate of $5e-5$. We adopt the AdamW~\cite{AdamW} optimizer with a constant weight decay of $0.1$ throughout the training. The gradient clip is set to $1.0$. We conduct gradient accumulation with an equivalent batch size of $64$ and train using \textit{bfloat16} precision. The training process consumes around $262$ GPU Hours (A100-40G). 

\subsection{Benchmarks}
\myparagraph{ChartQA}~\cite{ChartQA} test split comprises $1,250$ questions from both human-generated and augmented segments. The charts are sourced from web crawls with three prevalent chart types. ChartQA requires the model to respond to questions with only a single word or phrase and employs a lenient matching method to verify the correctness of the answers. Considering the impact of inference length on performance, instead of prompting the model to produce the shortest possible answers, we adopt a template-based answer extraction method, i.e., \textit{provide your final answer in} \textbackslash$box\{\}$. Refer to Appendix~\ref{apdx_prompt} for details. This approach effectively enhances the performance of mainstream models.

\myparagraph{ChartBench}~\cite{ChartBench} offers charts that lack data point annotations. It encompasses $9$ main categories and $42$ subcategories, with each sub-category housing $50$ charts. ChartBench emphasizes the reliability of chart numbers, presenting a stiffer challenge since models are unable to obtain precise answers via OCR. The models must understand each element of the chart to estimate values close to the ground truth. This benchmark uses a relaxed accuracy similar to ChartQA, and we also adopt the inference prompt of template extraction to boost model performance.

\subsection{Comparative Models}
We divide all methods into two groups: general MLLMs and those specifically designed for chart understanding. 

\myparagraph{General MLLMs}. We compare LLaVA-v1.5~\cite{llava15}, which paved the way for image-text interaction through visual instruction fine-tuning. We also compare the QwenVL series, including v1~\cite{qwen-vl}, v2~\cite{Qwen2VL}, and v2.5~\cite{Qwen25VL}. Due to its strong base performance, we set this series as the baseline for our {\Model}. We select Phi-3.5-Vision~\cite{phi3}, which is easy to deploy on the edge devices, and the Intern series for their high performance, such as InternlmXComposer-v2~\cite{internlm-xcomposerv2} and InternVL-v2.5~\cite{InternVL25}. We also provide the result of DeepSeekVL2~\cite{deepseekv2}, which is based on the MoE architecture. Note that we chose the versions of these models at around 10B for fair comparisons.

\myparagraph{Specialist chart models}. We provide classic chart methods like Matcha~\cite{MatCha} and Deplot~\cite{DePlot}. However, we adopt LLaVA-v1.6~\cite{llava16} to further analyze and summarize their output for meaningful comparisons. We also compare ChartVLM~\cite{ChartVLM}, ChartAst~\cite{ChartAst}, DocOwl-v1.5~\cite{DocOwl}, OneChart~\cite{OneChart}, and ChartLLama~\cite{Chartllama}, which are fine-tuned with chart-specific instructions. Since the Program of thought (PoT) can effectively improve the numerical calculation ability of MLLMs, we select ChartGemma~\cite{chartgemma}, TinyChart~\cite{TinyChart}, and ChartMoE~\cite{ChartMoE} for comparisons.

\subsection{Comparison with SOTA}

\myparagraph{Comparisons on ChartQA.}
Tab.~\ref{tab_chartqa} presents the performance of {\Model} on ChartQA. We report the relaxed accuracy for three different margins and provide detailed results for two distinct parts. {\Model} significantly outperforms the baselines, e.g., {\Mone} 83.16\%~\cite{Qwen2VL} vs. 85.28\% (+2.12\%$\uparrow$) and {\Mtwo} 86.36\%~\cite{Qwen25VL} vs. 87.74\% (+1.38\%$\uparrow$). Even though the Qwen-VL series models demonstrate sufficiently high baseline performance, {\Model} still manages to achieve remarkable enhancements, especially in the challenging Human-annotated part. This indicates that point-based CoT training can significantly improve the model's ability to read and understand charts. Notably, {\Model} also outperforms PoT-based methods~\cite{chartgemma, TinyChart, ChartMoE}. For example, when compared with ChartMoE+PoT~\cite{ChartMoE}, {\Model} attains 84.64\% vs. 87.74\% (+3.10\%$\uparrow$). This implies that increasing the reasoning length contributes to enhancing the model's numerical and logical capabilities, effectively overcoming scenarios involving extensive numerical calculations. 

\myparagraph{Comparisons on ChartBench.}
Tab.~\ref{tab_chartbench} shows the performance of {\Model} on ChartBench, where we report the detailed performance across 9 types of charts. Compared to ChartQA, {\Model} demonstrates more significant improvements on ChartBench, e.g., {\Mone} 58.90\%~\cite{Qwen2VL} vs. 62.61\% (+3.71\%$\uparrow$) and {\Mtwo} 60.91\%~\cite{Qwen25VL} vs. 65.95\% (+5.04\%$\uparrow$). While better OCR capabilities can enhance model performance on ChartQA, ChartBench focuses on data points without text annotations, which benefits more from superior chart element localization and reasoning abilities. This supports the advantage of point-based CoT over text-only CoT. Specifically, the improvement is more significant on \textit{extra} type charts, e.g., {\Mtwo} 57.26\%~\cite{Qwen25VL} vs. 65.03\% (+7.77\%$\uparrow$). This suggests that Point-based CoT training enables the model to develop a logical chart-reading process and comprehension skills, enhancing its generalization even to uncommon chart types.

%% file: tables/tab_chartbench.tex
\begin{table*}[t]
\centering
\captionsetup{skip=3pt}
\caption{The accuracy (\%) performance on \textbf{\textit{ChartBench}}. Our proposed {\Model} consistently outperforms other MLLMs remarkably.}
\resizebox{\linewidth}{!}{
\setlength{\tabcolsep}{12pt}
\begin{tabular}{@{}ccccccccccccc@{}}
\toprule[1pt]
\multicolumn{1}{c|}{\multirow{2}{*}{Models}} & \multicolumn{4}{c|}{Regular Type} & \multicolumn{7}{c|}{Extra Type} & \multirow{2}{*}{ALL} \\ \cmidrule(lr){2-12}
\multicolumn{1}{c|}{} & Line & Bar & Pie & \multicolumn{1}{c|}{Avg.} & Area & Box & Radar & Scatter & Node & Combin. & \multicolumn{1}{c|}{Avg.} &  \\ \midrule
\multicolumn{13}{c}{\textit{General MLLMs}} \\
\multicolumn{1}{c|}{LLaVA-v1.5~\cite{llava15}} & 29.12 & 21.26 & 17.28 & \multicolumn{1}{c|}{22.10} & 21.73 & 20.94 & 27.50 & 23.47 & 36.80 & 24.30 & \multicolumn{1}{c|}{24.96} & 23.38 \\
\multicolumn{1}{c|}{Qwen-VL~\cite{qwen-vl}} & 38.00 & 20.71 & 38.24 & \multicolumn{1}{c|}{29.46} & 28.83 & 24.17 & 35.00 & 19.50 & 18.50 & 25.50 & \multicolumn{1}{c|}{26.56} & 28.18 \\
\multicolumn{1}{c|}{Mini-Gemini~\cite{minigemini}} & 34.88 & 36.12 & 40.40 & \multicolumn{1}{c|}{36.77} & 31.20 & 23.33 & 30.60 & 35.20 & 43.60 & 27.90 & \multicolumn{1}{c|}{30.61} & 34.37 \\
\multicolumn{1}{c|}{InternlmXC-v2~\cite{internlm-xcomposerv2}} & 68.16 & 48.74 & 56.60 & \multicolumn{1}{c|}{54.50} & 27.47 & 25.33 & 40.10 & 52.93 & 50.40 & 46.20 & \multicolumn{1}{c|}{39.72} & 48.41 \\
\multicolumn{1}{c|}{InternVL-v2.5~\cite{InternVL25}} & 75.20 & 48.31 & 52.00 & \multicolumn{1}{c|}{55.09} & 32.00 & 20.00 & 44.00 & 45.33 & 70.00 & 48.00 & \multicolumn{1}{c|}{42.11} & 49.43 \\
\multicolumn{1}{c|}{DeepSeekVL2~\cite{deepseekvl2}} & 69.28 & 49.66 & 47.40 & \multicolumn{1}{c|}{53.71} & 40.80 & 44.40 & 40.50 & 76.14 & 45.40 & 59.50 & \multicolumn{1}{c|}{51.31} & 53.02 \\
\multicolumn{1}{c|}{Qwen2-VL~\cite{Qwen2VL}} & 74.40 & 50.77 & 63.00 & \multicolumn{1}{c|}{58.36} & \textbf{56.93} & 40.00 & 50.00 & \textbf{81.33} & 64.00 & 68.00 & \multicolumn{1}{c|}{\textbf{59.40}} & 58.90 \\
\multicolumn{1}{c|}{Qwen2.5-VL~\cite{Qwen25VL}} & \textbf{80.88} & \textbf{54.06} & \textbf{68.20} & \multicolumn{1}{c|}{\textbf{62.73}} & 37.33 & \textbf{46.13} & \textbf{51.90} & 72.27 & \textbf{74.40} & \textbf{74.00} & \multicolumn{1}{c|}{57.26} & \textbf{60.91} \\ \midrule
\multicolumn{13}{c}{\textit{Specialist Chart Models}} \\
\multicolumn{1}{c|}{Matcha~\cite{MatCha}} & 6.80 & 5.05 & 3.60 & \multicolumn{1}{c|}{5.18} & 0.27 & 1.60 & 6.20 & 3.46 & 5.40 & 4.80 & \multicolumn{1}{c|}{5.81} & 4.84 \\
\multicolumn{1}{c|}{ChartVLM~\cite{ChartVLM}} & 21.92 & 14.16 & 10.50 & \multicolumn{1}{c|}{15.16} & 7.47 & 7.87 & 8.00 & 7.87 & 5.40 & 10.50 & \multicolumn{1}{c|}{8.38} & 11.96 \\
\multicolumn{1}{c|}{ChartLlama~\cite{Chartllama}} & 26.80 & 18.83 & 20.80 & \multicolumn{1}{c|}{20.99} & 14.27 & 12.00 & 24.30 & 27.73 & 26.20 & 25.80 & \multicolumn{1}{c|}{21.71} & 21.31 \\
\multicolumn{1}{c|}{TinyChart~\cite{TinyChart}} & 32.40 & 25.81 & 22.50 & \multicolumn{1}{c|}{26.71} & 10.13 & 14.80 & 13.40 & 28.14 & 10.80 & 21.60 & \multicolumn{1}{c|}{22.56} & 22.51 \\
\multicolumn{1}{c|}{Deplot~\cite{DePlot}} & 31.20 & 26.46 & 24.00 & \multicolumn{1}{c|}{27.09} & 21.34 & 13.34 & 24.00 & 41.34 & 42.00 & 31.00 & \multicolumn{1}{c|}{31.57} & 27.62 \\
\multicolumn{1}{c|}{OneChart~\cite{OneChart}} & 41.28 & 30.28 & 29.60 & \multicolumn{1}{c|}{32.65} & 19.07 & 13.20 & 24.60 & 38.53 & 34.80 & 27.90 & \multicolumn{1}{c|}{31.91} & 29.93 \\
\multicolumn{1}{c|}{DocOwl-v1.5~\cite{DocOwl}} & 49.60 & 31.69 & 31.54 & \multicolumn{1}{c|}{35.68} & 12.27 & 23.33 & 22.50 & 36.13 & 29.60 & 38.80 & \multicolumn{1}{c|}{27.38} & 32.05 \\
\multicolumn{1}{c|}{ChartGemma~\cite{chartgemma}} & 50.48 & 38.21 & 32.10 & \multicolumn{1}{c|}{39.89} & 28.27 & 24.13 & 28.10 & 48.00 & 41.80 & 43.40 & \multicolumn{1}{c|}{42.47} & 38.46 \\
\multicolumn{1}{c|}{ChartMoE~\cite{ChartMoE}} & 71.44 & 51.57 & 52.80 & \multicolumn{1}{c|}{56.31} & 38.40 & 24.13 & 40.20 & 62.67 & 58.00 & 49.20 & \multicolumn{1}{c|}{55.58} & 51.67 \\
\rowcolor{blue!5}\multicolumn{1}{c|}{{\Mone}} & 79.84 & 54.58 & 68.24 & \multicolumn{1}{c|}{63.04} & \textbf{58.20} & 44.12 & 52.40 & \textbf{83.67} & 68.24 & 68.92 & \multicolumn{1}{c|}{62.09} & 62.61 \\
\rowcolor{blue!5}\multicolumn{1}{c|}{{\Mtwo}} & \textbf{82.40} & \textbf{58.88} & \textbf{71.40} & \multicolumn{1}{c|}{\textbf{66.71}} & 51.44 & \textbf{48.33} & \textbf{56.90} & 77.27 & \textbf{78.00} & \textbf{80.20} & \multicolumn{1}{c|}{\textbf{65.03}} & \textbf{65.95} \\ \bottomrule[1pt]
\end{tabular}
}
\label{tab_chartbench}
\vspace{-18pt}
\end{table*}

%% file: sec/5_further_discussion.tex
\section{In-depth Analysis}

\input{tables/tab_ablation}
\input{tables/tab_backbone}

\subsection{Ablation on Training Recipe}
Tab.~\ref{tab_ablation} presents the ablation study on our training recipe. As shown in Tab.~\ref{tab_sft_data}, we conduct the high-quality chart knowledge alignment before instruction tuning (+Stage1). We design detailed reasoning steps based on advanced LLMs (Fig.~\ref{fig_pipeline_cot}) to ensure even smaller models ($\sim$7B) can also benefit from inference scaling laws (+CoT). Additionally, we integrate grounding supervision into the CoT steps, enabling the model to continuously reflect on its reasoning and interact with input charts to refine the reasoning chain (+PointCoT). Since the baseline model is optimized for ChartQA during pre-training, the Stage1 alignment training yields marginal performance improvements (e.g., Qwen2-VL +0.58\%$\uparrow$, Qwen2.5-VL +0.14\%$\uparrow$). Direct distillation from reasoning steps also shows limited improvement because: 1) In Fig.~\ref{fig_pipeline_cot}, we adopt the LLM (not MLLM), so the reasoning process does not leverage chart information; 2) both ChartQA and ChartBench focus more on data point accuracy rather than numerical calculation or reasoning. Hence, textual CoT does not improve the model's accuracy in reading basic numbers from the chart. With grounding supervision, the model performance gets significantly improved, particularly on sparse-annotated ChartBench (Qwen2-VL +3.71\%$\uparrow$, Qwen2.5-VL +4.28\%$\uparrow$).

\input{tables/tab_bbox_format}
\input{tables/tab_abl_PE}

\input{figs/texs/visualize}

\subsection{Ablation on Backbone}
To demonstrate the effect of MLLMs for SFT based on PointCoT, we select two baseline models with relatively poor localization but strong chart-processing abilities for comparisons. As shown in Tab.~\ref{tab_backbone}, PointCoT is highly dependent on the underlying localization capabilities. Although both Qwen-VL~\cite{qwen-vl} and ChartMoE~\cite{ChartMoE} perform excellently in handling chart data, the reflection based on BBox fails to enhance their performance further. In contrast, both Qwen2-VL~\cite{Qwen2VL} and Qwen2.5-VL~\cite{Qwen25VL} can accurately indicate the objects using either points or BBoxes. Correspondingly, this enables PointCoT to work effectively, achieving a performance improvement of more than 1\%. 

\subsection{Ablation on Bounding Box Format}
Our proposed {\Model} reflects on the chart regions by outputting $(X_\text{top left}, Y_\text{top left}), (X_\text{bottom right}, Y_\text{bottom right})$ as bounding boxes. Our observations reveal that the numerical representation format significantly impacts the tuning process.  Table~\ref{tab_bbox_format} presents three formats using baselines trained on {\Dataset} for one epoch without additional data or tricks. \textit{Type A} normalizes numbers to four-decimal values between 0 and 1, representing relative positions on the chart. However, it yields only a marginal performance improvement of 0.52\%.  \textit{Type B} rounds values to three decimal places. With the same data size and training time, it achieves a 1.26\% improvement, significantly outperforming type A. Further analysis suggests that Qwen’s tokenizer splits decimals into three-digit segments, potentially increasing token-level training difficulty for Type A.  \textit{Type C} retains the baseline positioning format, which varies across MLLMs, using numbers between 0 and 999 to represent relative positions. This approach proves particularly beneficial for grounding training, leading to a 1.68\% performance improvement in just one epoch.  These findings highlight the importance of numerical representation in optimizing model performance.  

\subsection{Ablation on Prompt Engineering}
To effectively utilize rule-based metrics for evaluation, researchers require models to respond with a direct number or phrase, i.e., \textit{direct} prompt. However, we observe that for models with excellent instruction-following capabilities, performance can be further improved by extending the reasoning length. This conclusion is well-established in reasoning models~\cite{OpenAIO1, DeepseekR1}. Still, it also applies to MLLMs that are not explicitly designed for reasoning, particularly when compared to prompts that generate only a single word. Tab.~\ref{tab_abl_PE} illustrates two types of PE on both the baseline and our {\Mone}, with modifications applied exclusively to the reasoning prompt while keeping the model parameters unchanged. For Qwen2-VL, adjusting the PE results in a 0.92\% performance improvement, particularly on the more challenging Human subset. Although {\Mone} already demonstrated strong performance, the PE provides an additional 0.55\% gain on ChartQA.

\subsection{Case Visualization}
Fig.~\ref{fig_visualize} demonstrates specific cases from ChartQA and ChartBench. We choose the powerful Qwen2.5-VL-72B~\cite{Qwen25VL} and GPT-4O~\cite{gpt4o} for comparison with our {\Mtwo}. We request the models to output BBox when generating the CoT steps to support their reasoning (Appendix~\ref{prompt_match}). As shown in Fig.~\ref{fig_visualize}, only our {\Mtwo} provide the BBoxes as required by the prompt, yielding more accurate numbers on charts with sparse text annotations.  

%% file: tables/tab_ablation.tex
\begin{table}[t]
\centering
\captionsetup{skip=3pt}
\caption{Ablation study of training data in Tab.~\ref{tab_sft_data}. CoT: stage 2 adopts the CoT data generated by Fig.~\ref{fig_pipeline_cot}. PointCoT: stage 2 adopts {\Dataset}.}
\resizebox{\linewidth}{!}{
\setlength{\tabcolsep}{5pt}
\begin{tabular}{@{}l|ccc|ccc@{}}
\toprule[1pt]
\multicolumn{1}{c|}{\multirow{2}{*}{Settings}} & \multicolumn{3}{c|}{ChartQA} & \multicolumn{3}{c}{ChartBench} \\ \cmidrule(l){2-7} 
\multicolumn{1}{c|}{}                          & Human   & Aug.   & Avg.    & Regular    & Extra   & Avg.     \\ \midrule
Qwen2-VL                                       & 72.08   & 94.24     & 83.16  & 58.36      & 59.40   & 58.90   \\
\quad +Stage1                                        & 72.76   & 94.72     & 83.74  & 60.62      & 60.12   & 60.39   \\
\quad +Stage1+CoT                                    & 73.58   & 94.64     & 84.11  & 60.94      & 60.54   & 60.76   \\
\rowcolor{blue!5}\quad +Stage1+PointCoT                               & 76.12   & 94.48     & 85.30  & 63.04      & 62.09   & 62.61   \\ \midrule
Qwen2.5-VL                                     & 78.96   & 93.80     & 86.38  & 62.73      & 58.93   & 61.67   \\
\quad +Stage1                                        & 79.16   & 93.88     & 86.52  & 64.22      & 60.82   & 62.68   \\
\quad +Stage1+CoT                                    & 79.76   & 93.52     & 86.64  & 64.48      & 61.16   & 62.98   \\
\rowcolor{blue!5}\quad +Stage1+PointCoT                               & 81.36   & 94.12     & 87.74  & 66.71      & 65.03   & 65.95   \\ \bottomrule[1pt]
\end{tabular}
}
\label{tab_ablation}
\vspace{-10pt}
\end{table}

%% file: tables/tab_backbone.tex
\begin{table}[t]
\centering
\captionsetup{skip=3pt}
\caption{Ablation study on different MLLMs. We report the average relax accuracy@0.05 on ChartQA and ChartBench. PointCoT: stage 2 adopts {\Dataset}.}
\resizebox{\linewidth}{!}{
\setlength{\tabcolsep}{15pt}
\begin{tabular}{@{}l|cc|cc@{}}
\toprule
\multicolumn{1}{c|}{Model} & ChartQA & $\Delta$    & ChartBench & $\Delta$       \\ \midrule
Qwen-VL~\cite{qwen-vl}     & 65.70        & -    & 28.18           & -      \\
\quad\quad +PointCoT            & 66.12        & \good{+0.42} & 27.92           & \bad{-0.26} \\
ChartMoE~\cite{ChartMoE}   & 81.20         & -    & 51.67           & -      \\
\quad\quad +PointCoT            & 81.36        & \good{+0.16} & 51.94           & \good{+0.27}   \\ \midrule
Qwen2-VL~\cite{Qwen2VL}    & 83.16        & -    & 58.90           & -      \\
\quad\quad +PointCoT            & 84.84        & \good{+1.68} & 62.12           & \good{+3.22}   \\
Qwen2.5-VL~\cite{Qwen25VL} & 86.36        & -    & 61.67           & -      \\
\quad\quad +PointCoT            & 87.48        & \good{+1.12} & 65.66           & \good{+3.99}   \\ \bottomrule
\end{tabular}
}
\label{tab_backbone}
\vspace{-15pt}
\end{table}

%% file: tables/tab_bbox_format.tex
\begin{table}[t]
\centering
\captionsetup{skip=3pt}
\caption{Ablation study of bounding box format on ChartQA. In the ground truth, we normalize the point number into 0-1 (retain 3/4 decimal) or 0-999 to indicate the grounding area.}
\resizebox{\linewidth}{!}{
\setlength{\tabcolsep}{3pt}
\begin{tabular}{@{}c|cc|cccccc@{}}
\toprule[1pt]
Settings & Normalize & Decimal & Human & $\Delta$ & Aug. & $\Delta$ & ALL & $\Delta$ \\ \midrule
Qwen2-VL & - & - & 72.08 & - & 94.24 & - & 83.16 & - \\
Type A & {[}0-1{]} & 4 & 73.52 & \good{+1.44} & 93.84 & \bad{-0.40} & 83.68 & \good{+0.52} \\
Type B & {[}0-1{]} & 3 & 74.68 & \good{+2.60} & 94.16 & \bad{-0.08} & 84.42 & \good{+1.26} \\
\rowcolor{blue!5}Type C & {[}0-999{]} & 0 & 75.36 & \good{+3.28} & 94.32 & \good{+0.08} & 84.84 & \good{+1.68} \\ \bottomrule[1pt]
\end{tabular}
}
\label{tab_bbox_format}
\vspace{-10pt}
\end{table}

%% file: tables/tab_abl_PE.tex
\begin{table}[t]
\centering
\captionsetup{skip=3pt}
\caption{Ablation study of prompt engineering (PE) on ChartQA. Direct: PE from ChartQA. Match: inference step by step and extract final answer via designed pattern.}
\resizebox{\linewidth}{!}{
\setlength{\tabcolsep}{8pt}
\begin{tabular}{@{}c|c|cccccc@{}}
\toprule[1pt]
Model                         & PE      & Human & $\Delta$    & Aug. & $\Delta$    & ALL   & $\Delta$    \\ \midrule
\multirow{2}{*}{Qwen2-VL}     & direct & 72.08 & -    & 94.24   & -    & 83.16 & -    \\
                              & match   & 73.84 & \good{+1.76} & 94.32   & \good{+0.08} & 84.08 & \good{+0.92} \\ \midrule
\multirow{2}{*}{{\Mone}}      & direct & 75.22 & -    & 94.24   & -    & 84.73 & -    \\
                              & match   & 76.12 & \good{+0.90}  & 94.48   & \good{+0.24} & 85.28 & \good{+0.55} \\ \bottomrule[1pt]
\end{tabular}
}
\label{tab_abl_PE}
\vspace{-15pt}
\end{table}

%% file: figs/texs/visualize.tex
\begin{figure*}[t!]
    \centering
      \begin{overpic}[width=1\linewidth, grid=False]{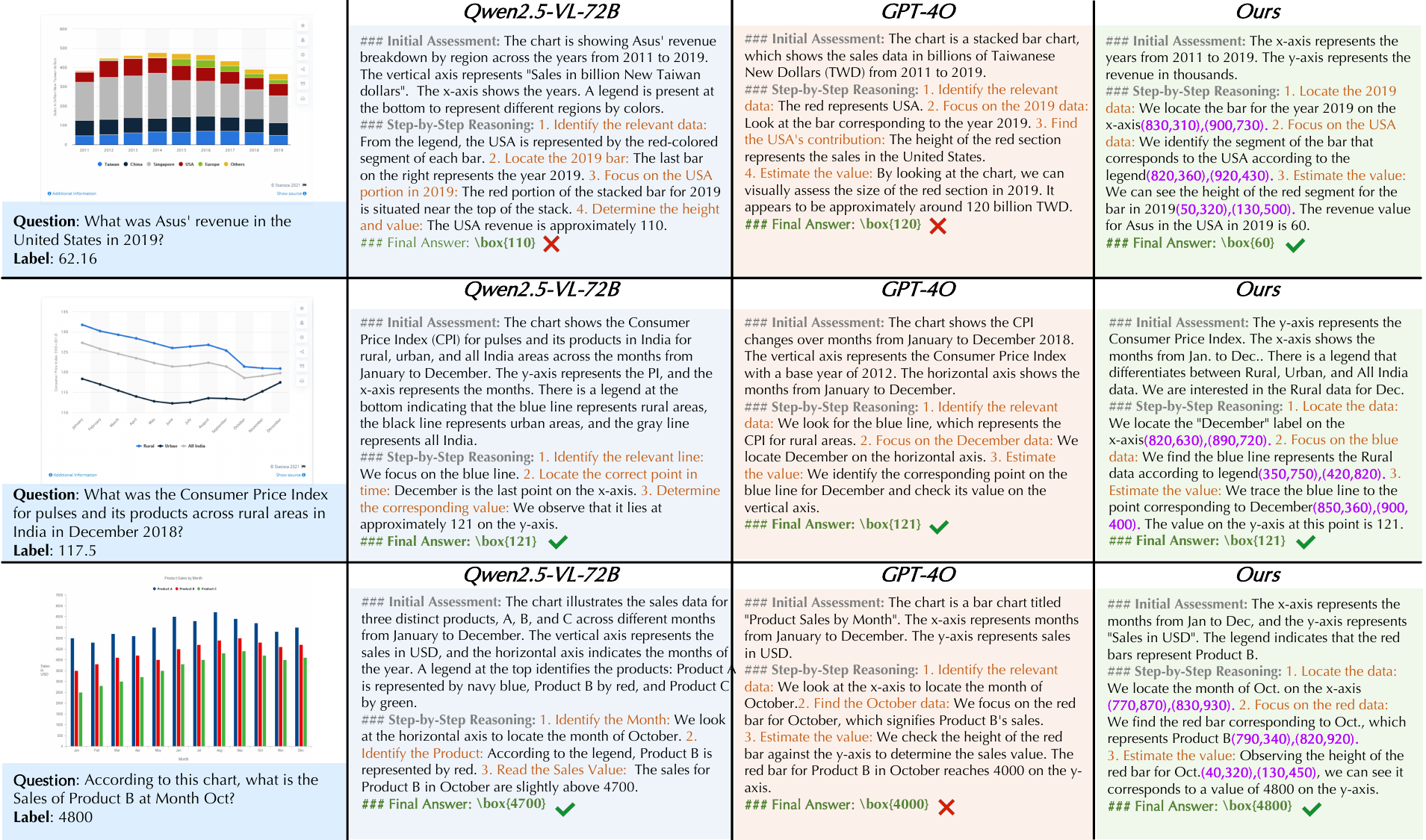}
      \end{overpic}
      \captionsetup{skip=3pt}   
      \caption{Comparsion between Qwen2.5-VL-72B~\cite{Qwen25VL}, GPT-4O~\cite{gpt4o} and {\Mtwo} (ours). All models adhere to the output format required by the prompt. However, both Qwen2.5-VL and GPT-4O ignore the BBox instruction. With the reflective output of the BBox, our {\Mtwo} has extracted precise numbers, and the BBoxes have provided sound explanations.}
      \vspace{-15pt}
    \label{fig_visualize}
\end{figure*}

%% file: sec/6_conclude.tex
\section{Conclusion}
We propose PointCoT, a multimodal CoT training method for chart understanding. We adopt the generated bounding boxes to verify whether the chain-of-thought reasoning steps are in line with the chart content. Specifically, we propose an automated annotation pipeline to provide the corresponding bounding boxes in the grounding steps and thus construct an instruction dataset. We provide two supervised fine-tuning models based on PointCoT data and conduct extensive experiments to demonstrate their effectiveness.  

%% file: sec/X_suppl.tex
\appendix
\onecolumn
\begin{center}
    \Large \textbf{ChartPoint: Guiding MLLMs with Grounding Reflection for Chart Reasoning}
    \Large \\ \textbf{Supplementary Material}
\end{center}
\vspace{20pt}

\section{Example Visualization}
\label{apdx_demo}
\input{figs/texs/apdx_case}

\clearpage

\section{Prompt Details}
\label{apdx_prompt}

\input{prompt/prompt_cot}
\input{prompt/prompt_code_edit}
\input{prompt/prompt_demo}

%% file: figs/texs/apdx_case.tex
\begin{figure}[h]
    \centering
      \begin{overpic}[width=0.95\linewidth, grid=False]{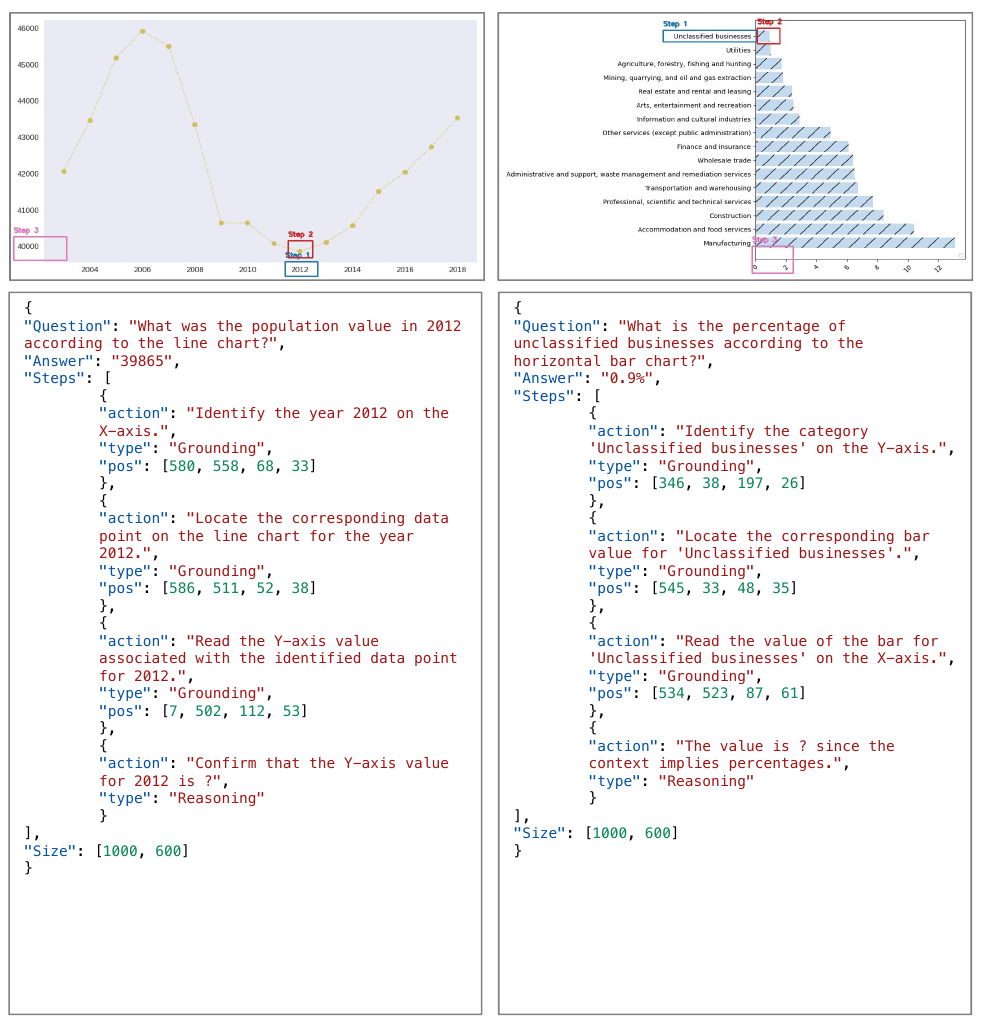}
      \end{overpic}
      \captionsetup{skip=3pt}      
      \caption{The detail of PointCoT and the display of the chart after BBox rendering (part 1).}
    \label{fig_apdx_case_1}
\end{figure}

\begin{figure}[h]
    \centering
      \begin{overpic}[width=0.95\linewidth, grid=False]{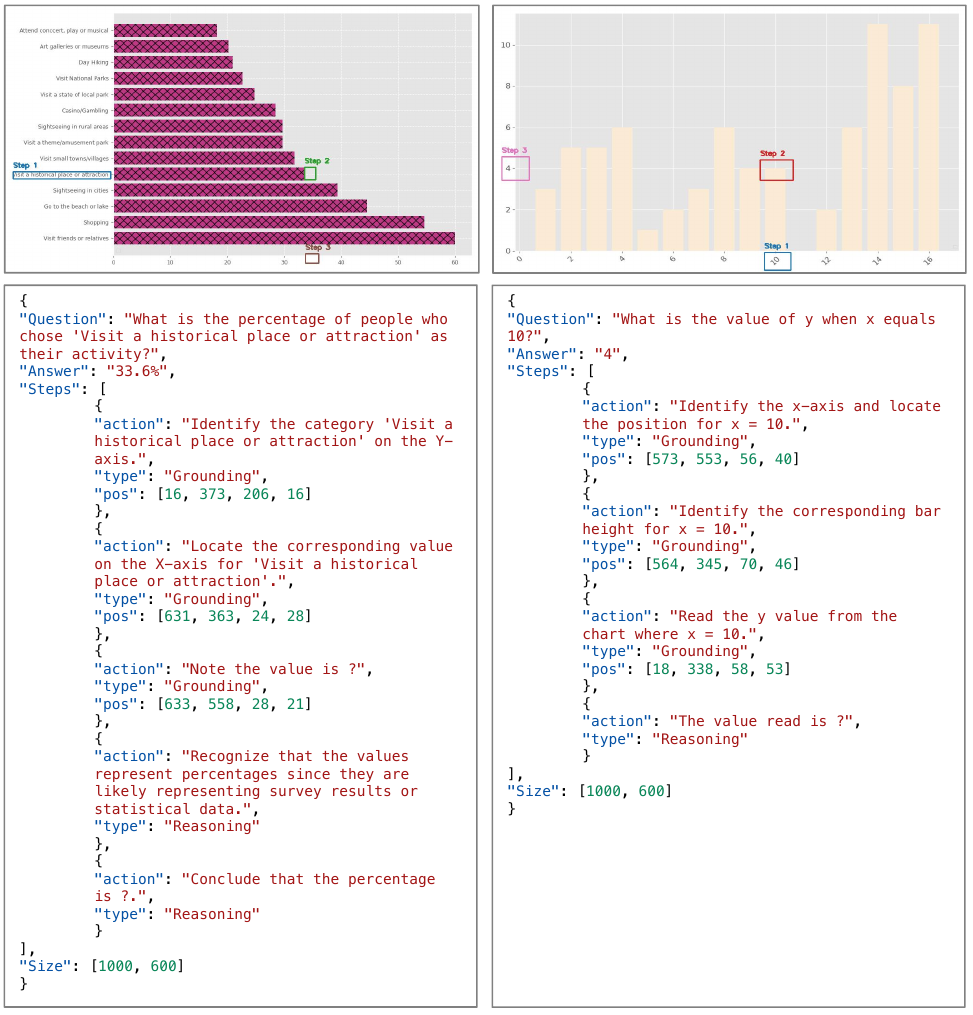}
      \end{overpic}
      \captionsetup{skip=3pt}      
      \caption{The detail of PointCoT and the display of the chart after BBox rendering (part 2).}
    \label{fig_apdx_case_2}
\end{figure}

%% file: prompt/prompt_cot.tex
\begin{tcolorbox}[colback=white,colframe=black!75,
    fonttitle=\bfseries, title=Prompt to Generate Chain of Thought,
    listing options={basicstyle=\ttfamily\small,breaklines=true},
    breakable] 
Given a chart plot code, please propose a \textbf{question} based on the code and provide the \textbf{answer} to this question. 

Additionally, provide detailed reasoning steps for arriving at the answer, categorizing each step as either {\color{lightBlue}\textbf{Grounding}} or {\color{dustyRose}\textbf{Reasoning}}.

\vspace{0.3cm}
{\color{darkNavy}\textbf{Requirements for the Question:}}
\begin{enumerate}
\item Focus on specific elements within the chart, such as the chart title, data values for specific categories, or identifying peak values.

\item {\color{dustyRose}\textbf{Do not}} ask descriptive or summary questions, like describing trends or summarizing the chart's message.

\item {\color{dustyRose}\textbf{Do not}} ask about visually non-distinguishing details, such as font type or size of the title.
\end{enumerate}
\vspace{0.3cm}
{\color{darkNavy}\textbf{Requirements for Thinking Steps:}}
\begin{enumerate}
\item Categorize each step into two types: {\color{lightBlue}\textbf{Grounding}} and {\color{dustyRose}\textbf{Reasoning}}.

\item {\color{lightBlue}\textbf{Grounding}} steps involve locating elements within the chart, such as identifying positions on axes or legend entries.

\item {\color{dustyRose}\textbf{Reasoning}} steps involve logical deductions based on information gathered from previous grounding steps.
\end{enumerate}
\vspace{0.3cm}
{\color{darkNavy}\textbf{Output Format Requirements:}}
\begin{enumerate}
\item Do not include any extraneous text unrelated to the content.

\item Strictly follow the JSON format for the response.

\item Refer to the example provided below for the expected structure.
\end{enumerate}
\vspace{0.3cm}
{\color{darkNavy}\textbf{Example:}}

\{\#EXAMPLE\_HERE\}

\vspace{0.3cm}
Now, given the chart plot code, please provide the \textbf{Question}, \textbf{Answer}, and \textbf{Steps}. 
Output strictly in the given example format.

\vspace{0.3cm}
{\color{darkNavy}\textbf{Chart Plot Code}}

\begin{lstlisting}[language=Python]
{#CODE_HERE}
\end{lstlisting}
Your should provide:
\label{prompt_cot}
\end{tcolorbox}

%% file: prompt/prompt_code_edit.tex
\begin{tcolorbox}[colback=white,colframe=black!75,
    fonttitle=\bfseries, title=Prompt for Code Editing,
    listing options={basicstyle=\ttfamily\small,breaklines=true},
    breakable] 
You are provided with Python code that generates a chart and a specific instruction related to an element within this chart that needs to be highlighted. Your task is to amend the plotting code by adding an `@' symbol at the location specified in the instruction.

\vspace{0.3cm}
{\color{darkNavy}\textbf{Requirements:}}
\begin{enumerate}
\item {\color{lightBlue}\textbf{Chart Plot Code}}: The original Python code for generating the chart.
\item {\color{dustyRose}\textbf{Instruction}}: A directive specifying which chart element should be highlighted (e.g., ``Locate the bar corresponding to `2018' and `Domestic'.", ``Circle the highest sales month").
\item \textbf{Modification}: Integrate an `@' symbol into the chart. This symbol should be centered on the element as indicated – for instance, in the middle of a title, legend, or label, or at the top of a bar.
\item \textbf{Output}: Deliver the updated code that correctly places the `@'.
\end{enumerate}

\vspace{0.3cm}
{\color{darkNavy}\textbf{Output Format:}}
\begin{enumerate}
\item Provide only the modified code as output.
\item Exclude any text not pertinent to the response content.
\item Ensure the modified code preserves all original chart features while incorporating the specified `@'.
\end{enumerate}

\vspace{0.3cm}
{\color{darkNavy}\textbf{Example:}}

\{\#EXAMPLE\_HERE\}

\vspace{0.3cm}
Now, given the {\color{lightBlue}\textbf{chart plot code}} and {\color{dustyRose}\textbf{instruction}}:

\vspace{0.3cm}
{\color{dustyRose}\textbf{Instruction}}: \{\#INST\_HERE\}.

\vspace{0.3cm}
{\color{lightBlue}\textbf{Chart Plot Code}}

\begin{lstlisting}[language=Python]
{#CODE_HERE}
\end{lstlisting}
Your should provide:
\end{tcolorbox}

%% file: prompt/prompt_demo.tex
\begin{tcolorbox}[colback=white,colframe=black!75,
    fonttitle=\bfseries, title=Match Style Prompt Engineering,
    listing options={basicstyle=\ttfamily\small,breaklines=true},
    breakable]
{\color{lightBlue}{\textbf{User}}}:

You are a professional data analyst with expertise in interpreting various types of charts and graphs. When presented with a question about a given chart, your response should adhere to the following guidelines.

\vspace{0.3cm}
{\color{darkNavy}\textbf{Guidelines:}}
\begin{enumerate}
\item \textbf{Initial Assessment}: Begin by carefully examining the provided chart. Note the type of chart (e.g., bar chart, pie chart, line graph), the labels on the axes (if applicable), the title, and any legends present. Identify the key data points relevant to the question.
\item \textbf{Step-by-Step Reasoning}: Break down the process of answering the question into clear, logical steps. Explain each step in detail, referencing specific data from the chart. Use phrases like ``First, we look at...", ``Next, we calculate...", ``Then, we compare...".
\item \textbf{Final Answer}: Conclude your response with the final answer presented in the format \textbackslash
box\{answer\}, where ``answer" is either a single number or a percentage. Do not include any units. Ensure that the answer is accurate based on your analysis.
\item \textbf{BBox}: Output bboxes whenever possible to support your understanding of the chart elements.
\end{enumerate}

\vspace{0.3cm}
{\color{dustyRose}\textbf{Question}}: \{\#QUESTION\_HERE\}.

\vspace{0.3cm}
{\color{lightBlue}{\textbf{Assistant}}}:
\label{prompt_match}
\end{tcolorbox}